\documentclass[runningheads]{llncs}
\usepackage{graphicx}
\usepackage{comment}
\usepackage{amsmath,amssymb} 
\usepackage{color}
\usepackage{xcolor}

\usepackage[width=122mm,left=12mm,paperwidth=146mm,height=193mm,top=12mm,paperheight=217mm]{geometry}

\usepackage{epsfig}
\usepackage{graphicx}
\usepackage{amsmath}
\usepackage{amssymb}
\usepackage[breaklinks=true,bookmarks=false]{hyperref}
\usepackage{color,soul}

\usepackage[labelformat=simple]{subcaption}

\usepackage{multirow}
\usepackage{cleveref}

\usepackage{booktabs}
\usepackage{url}
\hypersetup{
    colorlinks=true,
    linkcolor=blue,
    filecolor=magenta,      
    urlcolor=magenta,
}

\newcommand{\xhdr}[1]{\vspace{5pt} \noindent {\textbf{#1} }}
\newcommand{\etal}{\textit{et al}.}

\begin{document}

\pagestyle{headings}
\mainmatter

\title{When, Where, and What? A New Dataset 
for Anomaly Detection in Driving Videos}

\authorrunning{Yao~\etal}
\author{Yu Yao$^{1}$
\hspace{1cm} Xizi Wang$^2$
\hspace{1cm} Mingze Xu$^3\thanks{This work was done while the author was at Indiana University.}$
\hspace{1cm} Zelin Pu$^1$\\
\hspace{1cm} Ella M. Atkins$^{1}$
\hspace{1cm} David J. Crandall$^{2}$ 
\\
$^{1}$University of Michigan
\hspace{0.45cm} $^{2}$Indiana University
\hspace{0.45cm} $^{3}$Amazon Rekognition  \\
{\tt\small \{brianyao,ematkins\}@umich.edu, \{xiziwang,djcran\}@indiana.edu}
}
\institute{}

\titlerunning{When, Where, and What? 
Anomaly Detection in Driving Videos}

\maketitle

\begin{abstract}

Video anomaly detection (VAD) has been extensively studied. However, research on egocentric traffic videos with
dynamic scenes lacks large-scale benchmark datasets
as well as effective evaluation metrics.
This paper proposes traffic anomaly detection
with a \textit{when-where-what} pipeline to detect, localize,
and recognize anomalous events from egocentric videos.
We introduce a new dataset called Detection of Traffic Anomaly (DoTA) containing 4,677 videos with
temporal, spatial, and categorical annotations.  A new
spatial-temporal area under curve (STAUC) evaluation metric is proposed and used with DoTA.
State-of-the-art methods are benchmarked for two VAD-related tasks.
Experimental results show
STAUC is an effective VAD metric. To our knowledge, DoTA is the largest
traffic anomaly dataset to-date and is the first supporting traffic anomaly
studies across \textit{when-where-what} perspectives. Our code and dataset 
can be found in: \url{https://github.com/MoonBlvd/Detection-of-Traffic-Anomaly}
\end{abstract}
\vspace{-20pt}
\begin{figure}[htbp]
    \centering
    \includegraphics[width=0.9\textwidth]{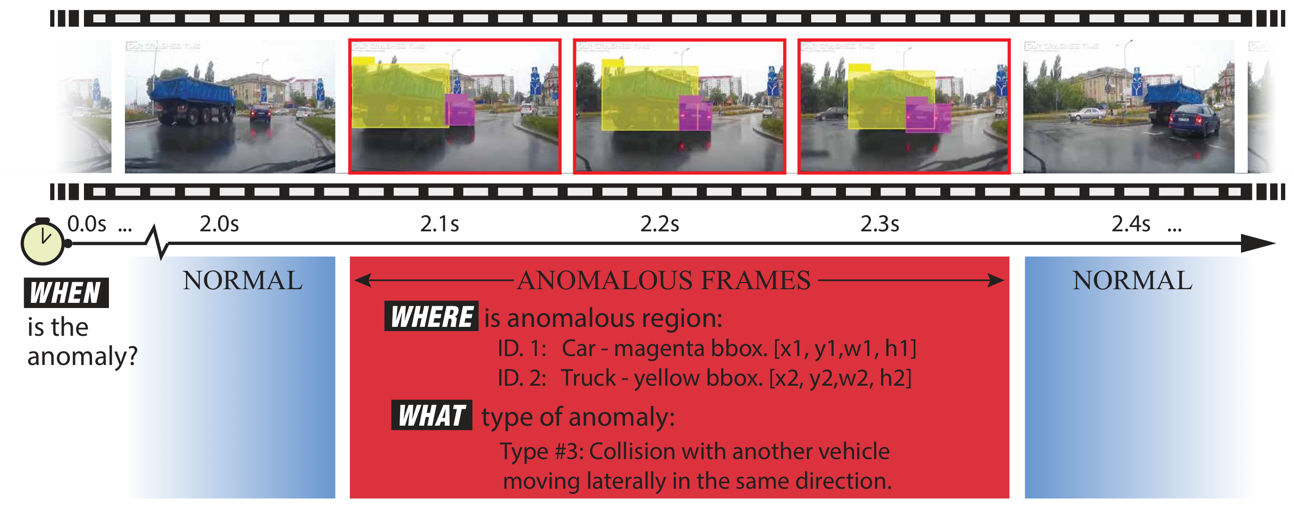}
    \vspace{-7pt}
    \caption{Overview of DoTA dataset. Annotations are provided to answer the \textbf{When}, \textbf{Where} and \textbf{What} questions for driving video anomalies.}
    \label{fig:teaser}
\end{figure}

\vspace{-30pt}
\section{Introduction}
Accurate perception of accident scenarios is a key challenge for advanced driver
assistance systems (ADAS).
Is an accident going to happen? Who will be involved? What type of accident is it?
These critical questions demand
detection, localization, and classification of on-road
anomalies for proper reaction and event data recording~\cite{yao2020sbb}.
We propose accident perception as a \textbf{When}-\textbf{Where}-\textbf{What}
pipeline:  \textbf{When} the anomalous event starts and ends, \textbf{Where} the anomalous regions are in each video frame,
and \textbf{What} the anomaly type is.
This pipeline can be mapped to two computer vision tasks:
video anomaly detection (VAD) and video action recognition (VAR).
VAD predicts per-frame anomaly scores to answer the \textbf{When} question, and computes per-pixel or per-object anomaly scores as intermediate step to implicitly answer the \textbf{Where} question. VAR classifies video type to answer the \textbf{What} question.

Training deep learning-based methods for VAD and VAR has been made possible by large-scale labeled datasets.
There are video datasets available for 
surveillance applications of VAD, including CUHK~\cite{lu2013abnormal}, 
ShanghaiTech Campus~\cite{liu2018future}, and UCF-crime~\cite{sultani2018real}, 
and for human activity recognition for VAR, including Sports-1M~\cite{KarpathyCVPR14} and Kinetics~\cite{kay2017kinetics}.
For traffic anomalies, recent first-person video datasets
such as 
StreetAccident~\cite{chan2016anticipating} and A3D~\cite{yao2019unsupervised}
have annotations of anomaly start and end times, while
DADA~\cite{fang2019dada} provides 
human attention maps from video spectator eye-gaze. However, no large-scale dataset 
and benchmark yet covers the full \textbf{When}-\textbf{Where}-\textbf{What} pipeline.

This paper introduces Detection of Traffic Anomaly (DoTA), a large-scale 
benchmark dataset for traffic VAD and VAR. DoTA contains  
$4,677$ videos with $18$ anomaly categories~\cite{bakker2017iglad}
and multiple anomaly participants in different driving scenarios. 
DoTA provides rich annotation for each anomaly: type (category),
temporal annotation, and anomalous object bounding box tracklets.
Taking advantage of this large-scale dataset with rich anomalous object annotations, 
we propose a novel VAD evaluation metric called Spatio-temporal Area 
Under Curve (STAUC). STAUC is motivated by the popular frame-level 
Area Under Curve (AUC). 
While AUC uses a per-frame anomaly score which is usually averaged from a pixel-level or object-level 
score map, 
STAUC takes such score map and computes how much of it overlaps with the annotated anomalous region. 
This overlap ratio is used as a weighting factor 
for true positive predictions with STAUC.  STAUC thus has AUC as its upper bound. 

We benchmark existing VAD baselines and state-of-the-art methods on DoTA using both 
AUC and STAUC. We also propose a simple-but-effective ensemble method 
that improves the performance of any single approach, offering a new direction to explore.
Extensive experiments show the importance of using this new metric in VAD research.
To further complete the pipeline, we also benchmark recent 
VAR methods such as R(2+1)D~\cite{tran2018closer} and SlowFast~\cite{feichtenhofer2019slowfast} 
on DoTA. Experiments show that applying generalized VAR methods to 
traffic anomaly understanding is far from perfect, motivating more research in this area.

This paper offers three contributions. First, we introduce DoTA, 
a large-scale ego-centric traffic video dataset to support VAD and VAR;
to the best of our knowledge this is the largest traffic video anomaly 
dataset and the first containing detailed temporal, spatial, and categorical annotations. Second, we identify problems with the commonly-used 
AUC metric and propose a new spatio-temporal evaluation metric (STAUC) to 
address them. We benchmark state-of-the-art VAD methods with both AUC and STAUC 
and show the effectiveness of our new metric.
Finally, we provide benchmarks of state-of-the-art VAR algorithms on 
DoTA, which we hope will encourage further research to manage challenging 
ego-centric traffic video scenarios. 
\section{Related Work}

\vspace{-5pt} \xhdr{Existing Video Anomaly Detection (VAD) datasets}
are generally from surveillance cameras. For example,
UCSD Ped1/Ped2~\cite{li2013anomaly},
CUHK Avenue~\cite{lu2013abnormal}, 
and ShanghaiTech~\cite{liu2018future}
were collected from campus surveillance cameras and
include
anomalies like prohibited objects and abnormal movements, while
UCF-Crime~\cite{sultani2018real} 
includes
accidents, robbery, and theft. 
Anomaly detection in egocentric traffic videos has very recently attracted attention. 
Chan~\etal~\cite{chan2016anticipating} propose the StreetAccident dataset
of on-road accidents with 620 video
clips collected from dash cameras. The last ten frames of each clip are annotated as anomalous.
Yao~\etal~\cite{yao2019unsupervised} propose the A3D dataset
containing 1,500 anomalous videos in which abnormal events are
annotated with the start and end times.
Fang~\etal~\cite{fang2019dada} introduce the DADA dataset
for driver attention prediction in accidents, while
Herzig~\etal~\cite{herzig2019spatio} extract a collision dataset
with 803 videos from BDD100K~\cite{yu2018bdd100k},
In contrast, our DoTA dataset is much larger (nearly 5,000) but, much 
more importantly, contains richer annotations
that support the whole
\textbf{When}-\textbf{Where}- \textbf{What} anomaly analysis pipeline.

\xhdr{Existing VAD models} 
mainly focus on the \textbf{When} problem
but are also implicitly related to \textbf{Where}.
Hasan~\etal~\cite{hasan2016learning} propose a convolutional Auto-Encoder (ConvAE)
to model the normality of video frames by reconstructing stacked input frames. 
Convolutional LSTM Auto-Encoder (ConvLSTMAE) is used
in~\cite{medel2016anomaly,chong2017abnormal,luo2017remembering}
capture regular visual and motion patterns.
Luo~\etal~\cite{luo2017revisit} propose a stacked RNN
for temporally-coherent sparse coding (TSC-sRNN).
Liu~\etal~\cite{liu2018future} detect anomalies by looking for differences
between predicted future frames and actual observations.
Gong~\etal~\cite{gong2019memorizing} propose an MemAE network to query pre-saved memory units for reconstruction, while
Wang~\etal~\cite{wang2019gods} design generalized one-class
sub-spaces for discriminative regularity modeling.
Other work has recently studied object-centric approaches.
Ionescu~\etal~\cite{ionescu2019object}
propose K-means to cluster object features
and train multiple support vector machine (SVM) classifiers with confidence as anomaly score.
Morais~\etal~\cite{morais2019learning} model human skeleton regularity
with local-global autoencoders and
compute per-object anomaly scores.
VAD in egocentric traffic scenarios is a new and challenging problem due
to dynamic foreground and background, perspective projection,
and complicated scenes. The most related work to ours is
TAD~\cite{yao2019unsupervised}, which predicts future object
bounding boxes from past time steps with RNN encoder-decoders, where
the standard deviation of predictions serves as the anomaly score. We benchmark stat-of-the-art VAD methods and their variants on DoTA dataset.

\xhdr{Action Recognition methods} address the
\textbf{What} problem to classify traffic anomalies.
Two-stream networks~\cite{simonyan2014two} and
temporal segment networks (TSN)~\cite{wang2016temporal}
leverage RGB and optical flow data.
Tran~\etal~\cite{tran2015learning} first proposed 3D convolutional networks (C3D) for
spatiotemporal modeling, followed by
an inflated model~\cite{carreira2017quo}. Recent work
substitutes 3D convolution with 2D and 1D convolution blocks (R(2+1)D~\cite{tran2018closer})
to improve effectiveness and efficiency.
Feichtenhofer~\etal~\cite{feichtenhofer2019slowfast} propose
the SlowFast model to extract video features
from low and high frame rate streams.
Online action detection in untrimmed, streaming videos
is addressed by De Geest~\etal~\cite{de2016online}, while
Gao~\etal~\cite{gao2017red} propose a reinforce encoder-decoder(RED)
to tackle action prediction and online action recognition.
Shou~\etal~\cite{shou2018online} model temporal consistency
with a generative adverserial network (GAN). Xu~\etal~\cite{xu2019temporal} propose
a temporal recurrent network (TRN)
leveraging future prediction to aid online action detection.
Gao~\etal~\cite{gao2019startnet} uses reinforcement learning
to detect the start time of actions. We benchmark VAR methods on DoTA dataset, and discuss online action detection in supplement.

\section{The Detection of Traffic Anomaly (DoTA) Dataset}\label{sec:dataset}

We introduce DoTA, the first publicly-available
\textbf{When}-\textbf{Where}-\textbf{What}
pipeline dataset with temporal, spatial, 
and categorical annotations.\footnote{The dataset will be made publicly available upon publication.}
To build DoTA, we collected more than 6,000 video clips from YouTube channels
and selected diverse dash camera accident videos from different 
countries under different weather and lighting conditions.  
We avoided videos with accidents that were not visible or camera fall-off from wind shield, 
resulting in 4,677 videos with $1280\times720$ resolution. 
Though the original videos are at $30$ fps, we extracted frames at $10$ fps 
for annotations and experiments in this paper. 
Table~\ref{tab:dataset} compares DoTA with other ego-centric traffic anomaly datasets. 

\begin{table}[t]
    \centering
    \caption{Comparison of published driving video anomaly datasets.}
    \label{tab:dataset}
    \scriptsize
    \renewcommand{\arraystretch}{1}
    \begin{tabular}{@{\;}l@{\quad}@{\quad}r@{\quad}@{\quad}rl@{\quad}@{\quad}l@{\;}}
        \toprule
        Dataset & \# videos & \multicolumn{2}{c}{\# frames} & Annotations \\
        \midrule
        StreetAccident~\cite{chan2016anticipating} & 620 & 62,000 &(20fps) & temporal \\
        A3D~\cite{yao2019unsupervised} & 1,500 & 128,175 &(10fps)  & temporal \\
        DADA~\cite{fang2019dada} & 2,000 & 648,476 &(30fps)& temporal, spatial (eye-gaze) \\
        \textbf{DoTA} & \textbf{4,677} & \textbf{731,932} &(10fps) & temporal, spatial (tracklets), categories \\
        \bottomrule
    \end{tabular}
    \vspace{-10pt}
\end{table}

We annotated the dataset using a custom tool based on 
Scalabel\footnote{https://scalabel.ai/}. 
Labeling traffic anomalies is subjective, especially for properties 
like start and end times.
To produce high quality annotations, each video was labeled by 
three annotators,
and the temporal and spatial (categorical) annotations were merged 
by taking average (mode) to minimize individual biases. 
Our 12 human annotators had different levels of driving experience.

\begin{table}[t]
    \centering
    \caption{Traffic anomaly categories in the DoTA dataset}
    \label{tab:anomaly_category}
    \scriptsize
    \begin{tabular}{@{\quad}c@{\quad}@{\quad}c@{\quad}@{\quad}l@{\;}}
        \toprule
         ID & Short & Anomaly Categories \\
         \midrule
         1  & ST & Collision with another vehicle which starts, stops, or is stationary\\
         2  & AH & Collision with another vehicle moving ahead or waiting\\
         3  & LA & Collision with another vehicle moving laterally in the same direction\\
         4  & OC & Collision with another oncoming vehicle\\
         5  & TC & Collision with another vehicle which turns into or crosses a road\\
         6  & VP & Collision between vehicle and pedestrian\\
         7  & VO & Collision with an obstacle in the roadway\\
         8  & OO & Out-of-control and leaving the roadway to the left or right \\
         9  & UK & Unknown \\
         \bottomrule
    \end{tabular}
\end{table}
\begin{figure}[t]
    \centering
    \includegraphics[width=0.9\columnwidth]{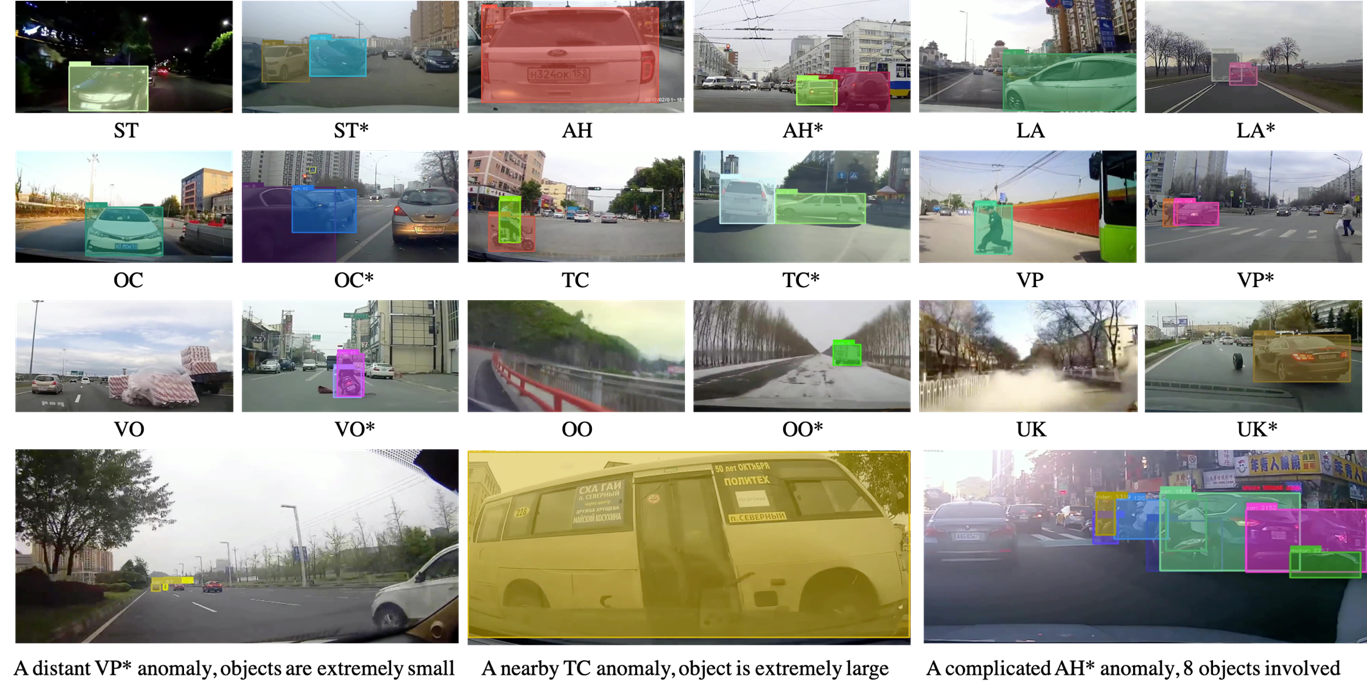}
    \caption{DoTA Samples. Spatial annotations are shown as shadowed bounding boxes. Short anomaly category labels with * indicate non-ego anomalies.}
    \label{fig:dota_sample}
    \vspace{-10pt}
\end{figure}

\begin{figure}[t]
    \begin{subfigure}[t]{0.48\textwidth}
        \centering
        \includegraphics[width=0.99\columnwidth]{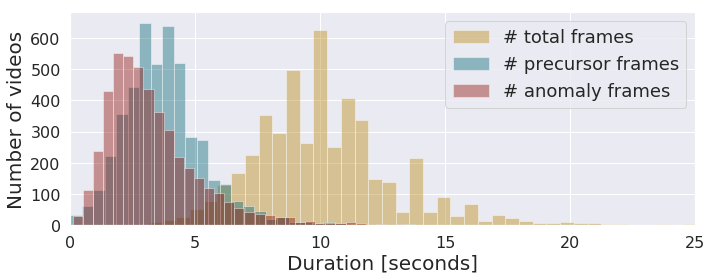}
        \vspace{-15pt}
        \caption{Duration distribution}
        \label{fig:durations}
    \end{subfigure}
    ~
    \begin{subfigure}[t]{0.48\textwidth}
        \centering
        \includegraphics[width=0.88\columnwidth]{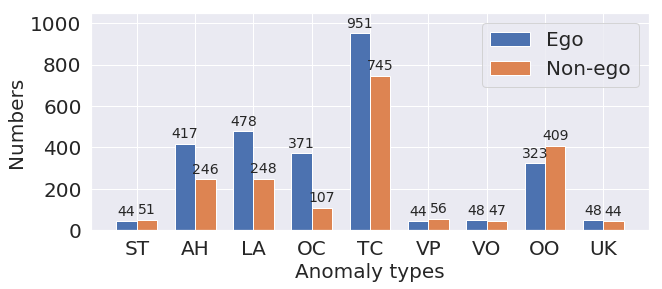}
        \vspace{-5pt}
        \caption{Anomaly category distribution}
        \label{fig:type_hist}
    \end{subfigure}
    
    \begin{subfigure}[t]{0.31\textwidth}
        \centering
        \includegraphics[width=0.95\columnwidth]{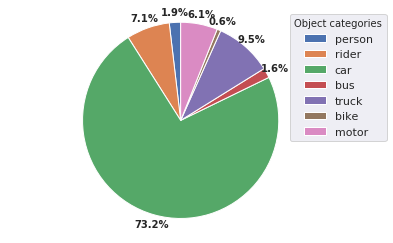}
        \vspace{-6pt}
        \caption{Object categories}
        \label{fig:obj_cat}
    \end{subfigure}
    ~
    \begin{subfigure}[t]{0.31\textwidth}
        \centering
        \includegraphics[width=0.95\columnwidth]{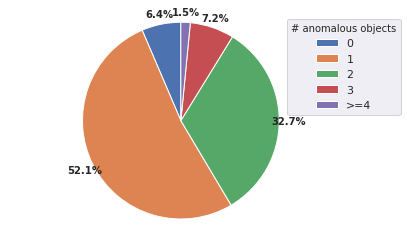}
        \vspace{-6pt}
        \caption{\# anomalous objects} 
        \label{fig:num_obj}
    \end{subfigure}
    ~
    \begin{subfigure}[t]{0.31\textwidth}
        \centering
        \includegraphics[width=0.95\columnwidth]{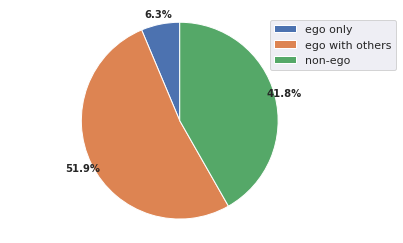}
        \vspace{-6pt}
        \caption{Ego-car involving}
        \label{fig:ego_involved}
    \end{subfigure}
    \caption{DoTA dataset statistics.}
    \label{fig:dataset_stats}
\end{figure}

\xhdr{Temporal Annotations.} Each DoTA video is annotated with anomaly start and end times, which separates 
it into three temporal
partitions: precursor, which is normal video preceding the anomaly, the anomaly window, and post-anomaly, which is normal activity following the anomaly.
Duration distributions are shown in Fig.~\ref{fig:durations}. 
Since early detection is essential for on-road 
anomalies~\cite{chan2016anticipating,suzuki2018anticipating}, 
we asked
the annotators to estimate the anomaly start as the time when the anomaly was inevitable.
The anomaly end was meant to be the time when all anomalous 
objects are out of the field of view or are stationary.
Our annotation is different from \cite{fang2019dada} where a frame is marked as anomaly 
start if half of the anomaly participant appears in the camera view; such a
start time can be too early because anomaly participants often
appear for a while before they start to behave abnormally. 
Our annotation is also distinct from 
\cite{chan2016anticipating} and \cite{yao2019unsupervised} 
where the anomaly start is marked when a crash happens, which does not support early detection.

\xhdr{Spatial Annotations.} DoTA is the first traffic anomaly dataset to provide 
detailed spatio-temporal annotation of anomalous objects. Each anomaly participant is assigned a unique
track ID, and their
bounding box is labeled from anomaly start to anomaly end or until the object is out of view.
We consider seven common traffic participant categories: 
person, car, truck, bus, motorcycle, bicycle, and rider,
following the BDD100K style~\cite{yu2018bdd100k}. Statistics of
object categories and per-video anomalous object numbers are 
shown in Fig.~\ref{fig:obj_cat} and~\ref{fig:num_obj}. 
DADA~\cite{fang2019dada} also provides spatial annotations
by capturing video observers' eye-gaze for driver attention studies. 
However, they have shown that 
eye-gaze does not always coincide with the anomalous region, and that
gaze can have $\sim$1 to 2 seconds  delay from anomaly start. 
Thus our tracklets provide improved annotation for spatio-temporal anomaly detection studies. 

\xhdr{Anomaly Categories.} Each DoTA video is assigned one of the 9 categories listed in Table~\ref{tab:anomaly_category}, as defined in  \cite{bakker2017iglad}.
We have observed that the same anomaly category with different viewpoints are visually distinct,
as shown in Fig.~\ref{fig:dota_sample}.
Therefore we split each category to ego-involved and non-ego (marked with *), 
resulting in 18 categories total.
Sometimes the category can be ambiguous, particularly when one anomaly is 
followed by another. For example, an oncoming out-of-control (OO*) vehicle 
might result in an oncoming collision (OC) with the ego vehicle. In such cases, 
we annotate the anomaly category as the dominant one in the video, 
i.e, the one that lasts longer during the anomaly period.
The distribution of videos of each category is shown in Fig.~\ref{fig:type_hist}.

\section{Video Anomaly Detection (VAD) Methods}\label{sec:benchmarked_methods}
We benchmark both unsupervised and supervised VAD. 
Unsupervised VAD is divided into frame-level 
and object-centric methods according to different input and output 
types. Supervised VAD is similar to temporal action detection but 
outputs a binary label indicating anomaly or no-anomaly.  

\vspace{-2pt}
\subsection{Frame-level Unsupervised VAD}
Frame-level unsupervised VAD methods detect anomalies
by either reconstructing past frames or predicting future frames 
and computing the reconstruction or prediction error. We benchmark three 
methods and their variants in this paper.
\begin{figure}[t]
    \begin{subfigure}[t]{0.18\textwidth}
        \centering
        \includegraphics[height=3cm]{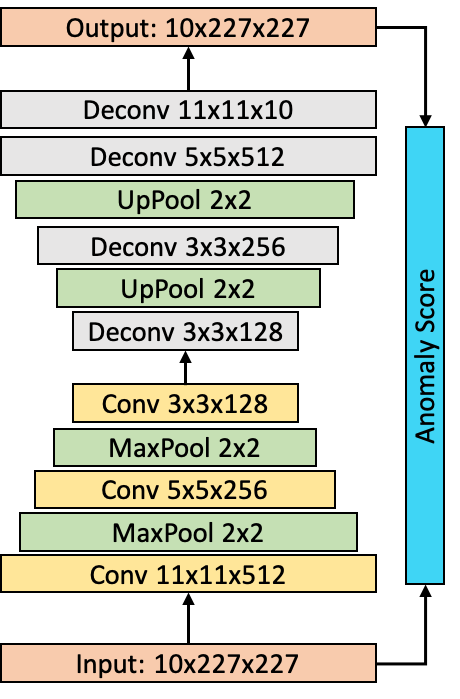}
        \captionsetup{font={scriptsize}}
        \caption{ConvAE} 
        \label{fig:ConvAE}
    \end{subfigure}
    ~
    \begin{subfigure}[t]{0.185\textwidth}
        \centering
        \includegraphics[height=3cm]{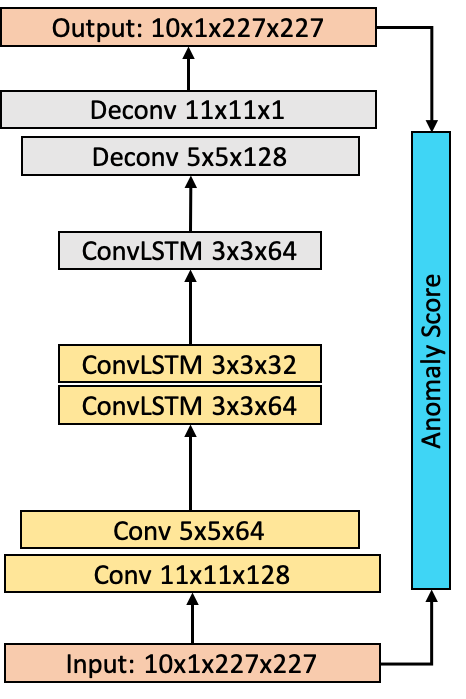}
        \captionsetup{font={scriptsize}}
        \caption{ConvLSTMAE}
        \label{fig:ConvLSTMAE}
    \end{subfigure}
    ~
    \begin{subfigure}[t]{0.18\textwidth}
        \centering
        \includegraphics[height=3cm]{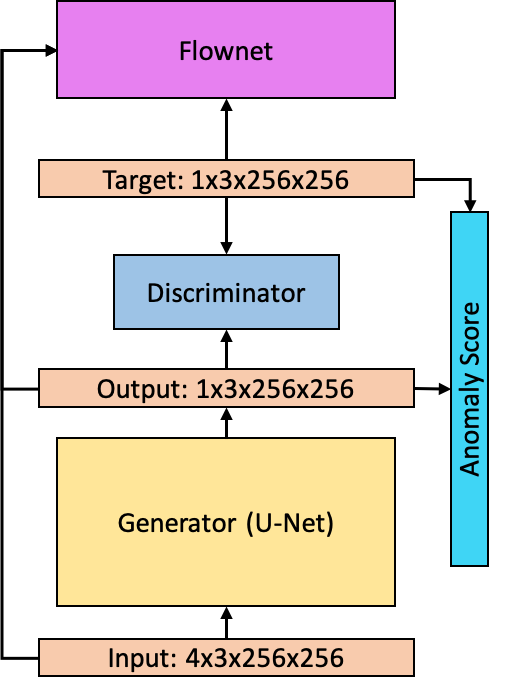}
        \captionsetup{font={scriptsize}}
        \caption{AnoPred} 
        \label{fig:AnoPred}
    \end{subfigure}
    ~
    \begin{subfigure}[t]{0.18\textwidth}
        \centering
        \includegraphics[height=3cm]{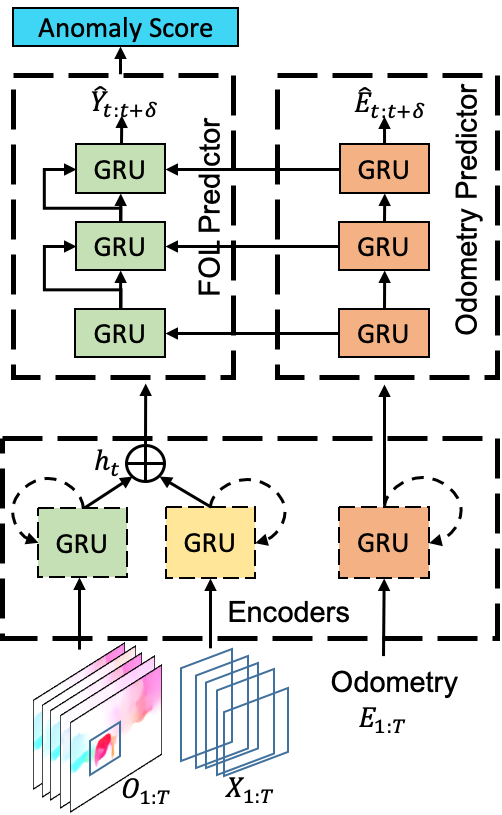}
        \captionsetup{font={scriptsize}}
        \caption{TAD} 
        \label{fig:TAD}
    \end{subfigure}
    ~
    \begin{subfigure}[t]{0.18\textwidth}
        \centering
        \includegraphics[height=3cm]{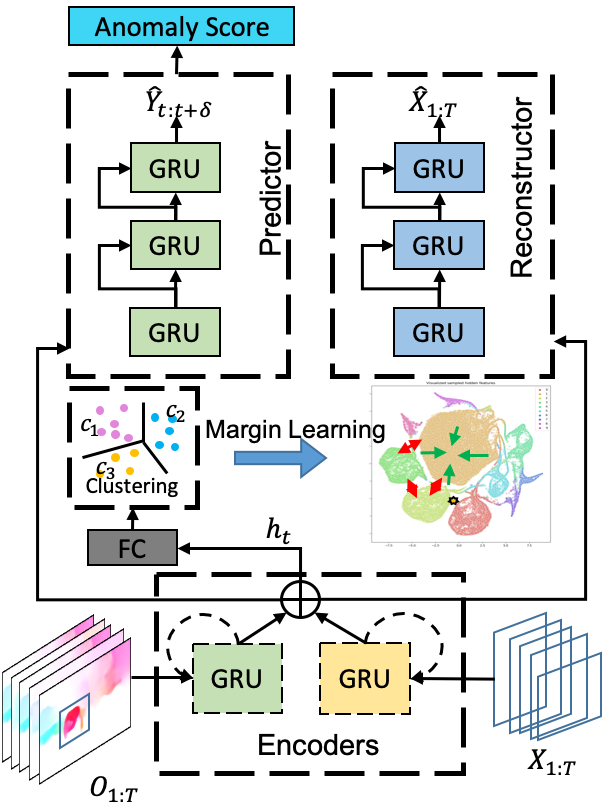}
        \captionsetup{font={scriptsize}}
        \caption{TAD+ML}
        \label{fig:TAD_ML}
    \end{subfigure}
    \caption{Network architecture of unsupervised VAD methods.}
    \label{fig:frame_VAD}
\end{figure}

\vspace{-4pt}
\xhdr{ConvAE}~\cite{hasan2016learning} is a 
spatio-temporal autoencoder model which encodes 
temporally stacked images with 2D convolutional 
encoders and decodes with deconvolutional 
layers to reconstruct the input 
(Fig.~\ref{fig:ConvAE}). The per-pixel 
reconstruction error forms an anomaly score 
map $\Delta I$ and the mean squared error (MSE)
is computed as a
frame-level anomaly score,
\begin{equation}
    MSE=\frac{1}{|M|}\sum_{i\in M}{\Delta I(i)}, \text{ and } \Delta I(i)=||I(i)-
    \hat{I}(i)||^2, \label{eq:mse_score}\\
\end{equation}
where $I$ and $\hat{I}$ are the ground truth 
and reconstructed/predicted frames, $M$ represents all frame pixels, and $\Delta I$ is 
also called anomaly score map.
To further compare 
the effectiveness of image  and motion 
features, we implement \textbf{ConvAE(gray)} 
and \textbf{ConvAE(flow)} to reconstruct 
the grayscale image and the dense optical 
flow, respectively. The input to ConvAE(flow) 
is a stacked historical flow map with 
size $20\times227\times227$, acquired from  
pre-trained FlowNet2~\cite{ilg2017flownet}.

\xhdr{ConvLSTMAE}~\cite{chong2017abnormal} is similar to ConvAE 
but models spatial and temporal features separately. 
A 2D CNN encoder first captures spatial information 
from each frame, then a multi-layer ConvLSTM 
recurrently encodes temporal features. Another 2D CNN decoder then reconstructs input video clips (Fig.~\ref{fig:ConvLSTMAE}). 
We also implemented \textbf{ConvLSTMAE(gray)} and \textbf{ConvLSTMAE(flow)}.

\xhdr{AnoPred}~\cite{liu2018future} is a frame-level VAD method taking four continuous previous RGB frames as input and 
applying UNet to predict a future RGB frame (Fig.~\ref{fig:AnoPred}). 
AnoPred boosts prediction accuracy with a multi-task loss incorporating
image intensity, optical flow, gradient, and adversarial losses.
AnoPred was proposed for surveillance cameras. However,
traffic videos are much more dynamic, making future frame prediction difficult. 
Therefore we also benchmark a variant of AnoPred to focus on video foreground. 
We use Mask-RCNN~\cite{he2017mask} 
pre-trained on Cityscapes~\cite{cordts2016cityscapes} 
to acquire object instance masks for each frame, and 
apply instance masks to input and target images, resulting in a 
\textbf{AnoPred+Mask} method that only predicts foreground objects and 
ignores noisy backgrounds such as trees and billboards. 
In contrast to~\cite{hasan2016learning,chong2017abnormal},
AnoPred uses Peak Signal to Noise Ratio, $PSNR = 10\log_{10} MSE^{-1}$ as anomaly score 
 with better results. 

\subsection{Object-centric Unsupervised VAD}

\xhdr{TAD}~\cite{yao2019unsupervised} models normal bounding box 
trajectories in traffic scenes with a multi-stream RNN encoder-decoder
~\cite{yao2019egocentric}
(Fig.~\ref{fig:TAD}) to encode past trajectories 
and ego motion and to predict future object bounding boxes. 
Prediction results are collected;  prediction consistency
instead of accuracy is used to compute per-object anomaly scores. 
Per-object scores are averaged to form a per-frame score.

Ionescu~\etal~\cite{ionescu2019object} propose to treat object normality as 
multi-modal and use k-means to find the normality clusters in hidden space.
Liu~\etal~\cite{liu2019margin} use margin learning (ML) to enforce large distances
between normal  and abnormal features. We combine these ideas and propose 
\textbf{TAD+ML}, as shown in Fig.~\ref{fig:TAD_ML}. We adopt k-means to 
cluster encoder hidden features. Each cluster is considered one normality, \textit{i.e.} one type of normal motion, 
so that each training sample is initialized with a cluster ID as its normality label.
Then we used a center loss~\cite{wen2016discriminative} to enforce 
tight distribution of samples from the same normality and to enforce 
samples from different normalities to be distinguishable. 
Center loss is more efficient than triplet loss~\cite{liu2019margin} 
in large batch training. Fig~\ref{fig:TAD_ML} shows an example of 
visualized hidden features after ML. 
Note that we removed the ego motion branch in TAD+ML for simplicity as it does not affect results.

\xhdr{Ensemble.} 
Frame-level VAD methods focus on appearance while object-centric methods 
focus more on object motion. We are not aware of any method combining 
the two. Appearance-only methods may fail with drastic variance 
in lighting conditions and motion-only
methods may fail when trajectory prediction is imperfect.
In this paper, we combine AnoPred+Mask and TAD+ML,
into an ensemble method.
We trained each method independently and fused 
their output anomaly scores by average pooling. 
We have observed that such a late fusion is better than fusing 
hidden features in an early stage and training the two models together, 
since their hidden features are scaled differently. AnoPred+Mask 
encodes one feature per frame, while TAD+ML has one feature per object. 

\subsection{Supervised VAD as Online Action Detection}

VAD can also be interpreted as binary action detection with normal and abnormal classes. 
We benchmark multiple video action 
detection methods on DoTA  to provide 
insight in supervised VAD. We use an 
ImageNet pre-trained ResNet50~\cite{he2016deep} 
model to collect frame features and train 
different classifiers: 1) \textbf{FC}, a 
three-layer fully-connected network for image 
classification;  2) \textbf{LSTM}, a one-layer 
LSTM classifier for sequential image 
classification; and 3) \textbf{Encoder-Decoder}, 
an LSTM model with an encoder classifying 
current frames and a decoder predicting future 
classes. We also train the temporal recurrent 
network (\textbf{TRN})~\cite{xu2019temporal} which is built upon
encoder-decoder except predictions 
are fed back to the 
encoder to improve performance.  

\section{A New Evaluation Metric}
\label{sec:metrics}

\subsection{Critique of Current VAD Evaluation}
\begin{figure}[t]
    \centering
    \begin{subfigure}[htbp]{0.19\textwidth}
        \centering
        \includegraphics[height=2.7cm]{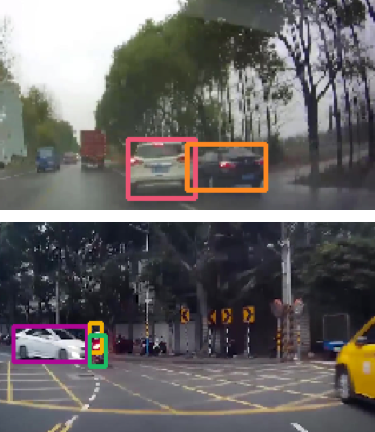}
        \captionsetup{font={scriptsize}}
        \caption{GT image}
        \label{fig:figure6a}
    \end{subfigure}
    \begin{subfigure}[htbp]{0.19\textwidth}
        \centering
        \includegraphics[height=2.7cm]{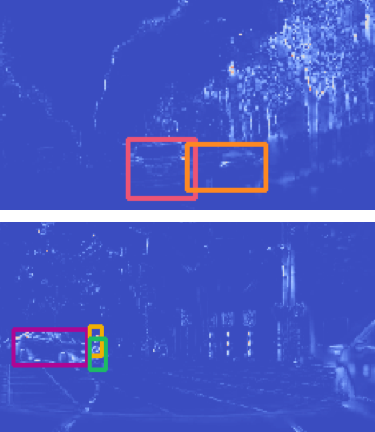}
        \captionsetup{font={scriptsize}}
        \caption{AnoPred}
        \label{fig:figure6b}
    \end{subfigure}
    \begin{subfigure}[htbp]{0.2\textwidth}
        \centering
        \includegraphics[height=2.7cm]{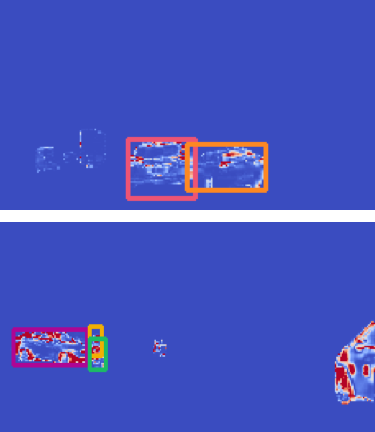}
        \captionsetup{font={scriptsize}}
        \caption{AnoPred+Mask}
        \label{fig:figure6c}
    \end{subfigure}
    \begin{subfigure}[htbp]{0.19\textwidth}
        \centering
        \includegraphics[height=2.7cm]{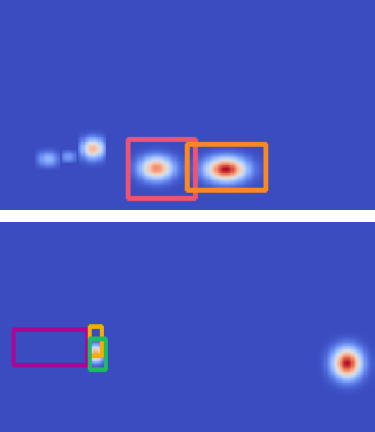}
        \captionsetup{font={scriptsize}}
        \caption{TAD+ML}
        \label{fig:figure6d}
    \end{subfigure}
    \begin{subfigure}[htbp]{0.19\textwidth}
        \centering
        \includegraphics[height=2.7cm]{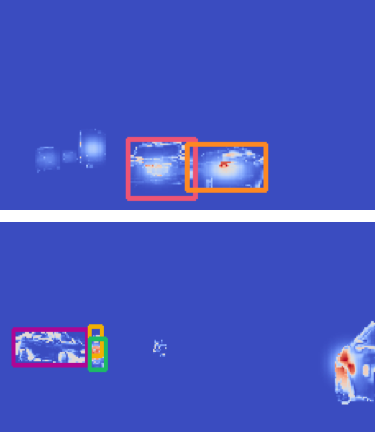}
        \captionsetup{font={scriptsize}}
        \caption{Ensemble}
        \label{fig:figure6e}
    \end{subfigure}
    \caption{Anomaly score maps computed by four methods. Ground truth anomalous regions are labeled by bounding boxes. Brighter color indicates higher score.} 
    \label{fig:false_true_positive}
\end{figure}

Most VAD methods compute an anomaly score for each frame by
averaging scores over all pixels or objects. Current evaluation 
method plots
receiver operating characteristic (ROC) curves using temporally 
concatenated scores and computes an area under curve (AUC) metric. 
AUC measures how well a VAD method answers the \textbf{When} question 
but ignores \textbf{Where}  since averaged anomaly score lacks 
spatial information. We argue AUC is insufficient to fully evaluate 
VAD performance. In computing AUC,
a true positive is a prediction where the model predicts high 
anomaly score for a positive frame.
Fig.~\ref{fig:false_true_positive} shows two positive frames and 
their
corresponding score maps computed by the four benchmarked VAD methods. 
Although the maps are different, the anomaly scores averaged from these maps
are similar, meaning they are treated similarly in AUC evaluation.  
This results in similar AUCs among all methods, which leads to a conclusion that all perform similarly. 
However, AnoPred (Fig.~\ref{fig:figure6b}) predicts high scores for 
trees and other noise. AnoPred+Mask and TAD+ML (Fig.~\ref{fig:figure6c} and~\ref{fig:figure6d})  
predict high scores for unrelated vehicles. Ensemble (Fig.~\ref{fig:figure6e}) alleviates
these problems but still has high anomaly scores outside the labeled 
anomalous regions. Note that score maps of TAD+ML and 
Ensemble are pseudo-maps introduced in Section~\ref{sec:stauc}.
Although these methods yield similar AUCs, VAD methods should be 
distinguished by their abilities to localize anomalous regions. 
Anomalous region localization is essential because it improves 
reaction to anomalies, e.g. collision avoidance, and aids in 
model explanation, e.g. a model predicts a car-to-car collision 
because it finds anomalous cars, not trees or noise. 
This motivates a new spatio-temporal metric to better address 
both \textbf{When} and \textbf{Where} questions.

\subsection{The Spatial-Temporal Area Under Curve (STAUC) Metric}\label{sec:stauc}
First, calculate the true anomalous region rate ($TARR$) for each positive frame, 
\begin{equation}
    TARR_t = \frac{\sum_{i\in m_t}\Delta I(i)}{\sum_{i\in M}\Delta I(i)}, \label{eq:tarr} 
\end{equation}
where $\Delta I$ is the anomaly score map from Eq.~\eqref{eq:mse_score}, $M$ represents all frame pixels, $m_t$ is the annotated anomalous frame region (i.e., 
the union of all annotated bounding boxes). $TARR\in[0,1]$ is a scalar describing 
how much of the anomaly score is located within the true anomalous region. 
$TARR$ is inspired by anomaly segmentation tasks where the overlap 
between prediction and annotation is computed~\cite{bergmann2019mvtec}. 
Next, calculate the spatio-temporal true positive rate ($STTPR$),
\begin{align}
    STTPR & = \frac{\sum_{t\in TP}{TARR_t}}{|P|}, \label{eq:sttpr}
\end{align}
where $TP$ represents all true positive predictions and $P$ represents
all ground truth positive frames. $STTPR$ is a weighted TPR 
where each true positive is weighted by its $TARR$. 
We then use $STTPR$ and FPR to plot a spatio-temporal ROC (STROC) 
curve and then calculate the STAUC. Note that  STAUC$\leq$AUC and the two are equal in the best case where $TARR_t=1$ $\forall t$.

Object-centric VAD~\cite{yao2019unsupervised,ionescu2019object,morais2019learning} 
computes per-object anomaly scores $s_k$ instead of an anomaly score map $\Delta I$. 
To generalize the STAUC metric to object-centric methods, 
we first create pseudo-anomaly score maps per Fig.~\ref{fig:figure6d}. 
Each object has a 2D Gaussian distribution centered in its bounding box.
Pixel score is then computed as the sum of the scores calculated from all boxes it occupies,
\begin{equation}
    \Delta I_{pseudo}(i)=\sum_{\forall k, i\in B_k} s_k\, e^{-\frac{|i_x-x_k|^2}{2w_k} - \frac{|i_y-y_k|^2}{2h_k}}, 
\end{equation}
where $i_x$ and $i_y$ are coordinates of pixel $i$ and $[x_k, y_k, w_k, h_k]$ is center location, width, and height of object bounding box $B_k$. For Ensemble method, we take the average of $\Delta I$ and $\Delta I_{pseudo}$ as the
anomaly score map in Fig.~\ref{fig:figure6e}. This map is used like $\Delta I$ in Eq.~\eqref{eq:tarr} to compute $TARR$ and STAUC.

$TARR$ is not robust to anomalous region size $m_t$. When $m_t\ll M$,  $TARR$ could be  small even though all anomaly scores are high in $m_t$. 
We thus propose selecting the top $N\%$ of pixels with the largest
anomaly scores as candidates, and compute $TARR$ from these 
candidates instead 
of all pixels. Selecting a constant $N$ can be arbitrary. An 
extremely small 
$N$ such as $0.01$ may result in a biased candidate set dominated by 
false or true detections such that $TARR=0$ or $1$. To address this 
issue, we compute an
adaptive $N$ for each frame based on the size of its annotated 
anomalous region as given by
\begin{align}\label{eq:adaptive_N}
    N_{adaptive}=\frac{\text{number of pixels in annotated anomalous region}}{\text{Total number of pixels}}\times 100.
\end{align}
The average $N_{adaptive}$ of DoTA is $11.12$ with a standard 
deviation $13.09$. The minimum and maximum $N_{adaptive}$ values 
are $0.005$ and $95.8$,
showing extreme cases where the anomalous object is very small 
(far away) or large (nearby).

\begin{figure}[t]
    \centering
    \begin{subfigure}[htb]{0.48\textwidth}
        \centering
        \includegraphics[width=0.9\textwidth]{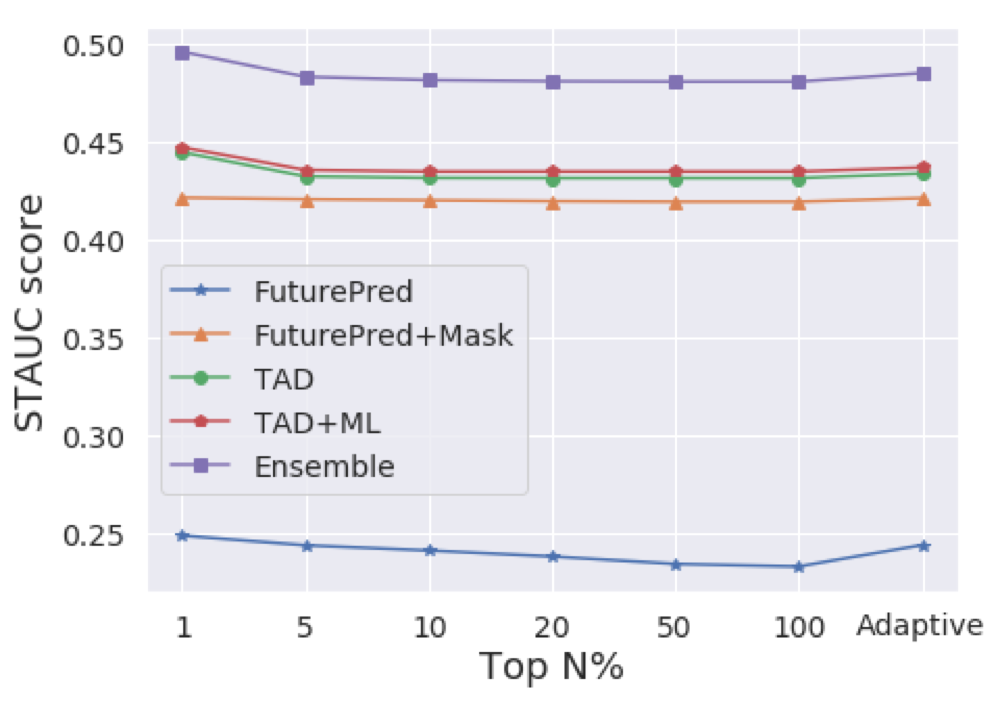}
        \captionsetup{font={scriptsize}}
        \caption{}
        \label{fig:stauc_robust}
    \end{subfigure}
    ~
    \begin{subfigure}[htb]{0.48\textwidth}
        \centering
        \includegraphics[width=0.9\textwidth]{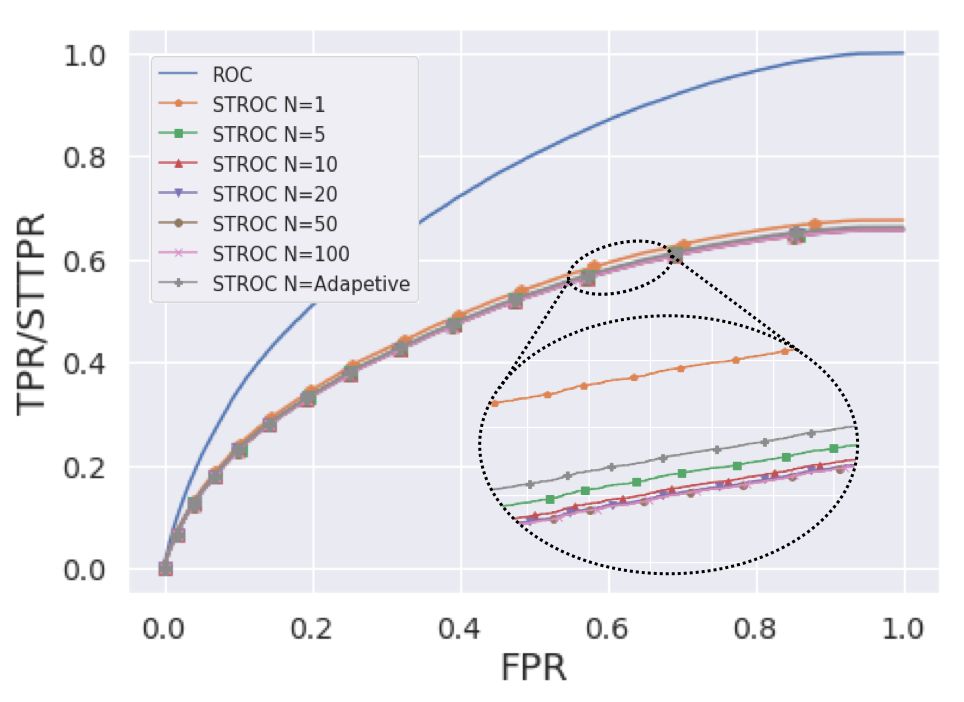}
        \captionsetup{font={scriptsize}}
        \caption{}
        \label{fig:stroc_robust}
    \end{subfigure}
    \caption{(a) STAUC values of different methods using different top $N\%$; (b) ROC curve and STROC curves of the Ensemble method with different top $N\%$.} 
\end{figure}

A critical consideration for any new metric is its robustness to  hyper parameters.
We have tested STAUC with $N=[1,5,10,20,50,100,N_{adaptive}]$ for different VAD 
methods per Fig.~\ref{fig:stauc_robust},  STAUC  slightly decreases with $N$ 
increasing but stabilizes when $N$ is large indicating STAUC is robust. Fig.~\ref{fig:stroc_robust} shows that STROC curves with different $N$ are close, especially when $N\geq5$, and their upper bound is the traditional ROC.  $N_{adaptive}$ is selected for our benchmarks based on each frame's annotation and its corresponding mid-range STAUC value.

\section{Experiments}\label{sec:experiment}
We benchmarked VAD and VAR with the \textbf{When}-\textbf{Where}-\textbf{What} pipeline. 
We randomly partitioned DoTA into 3,275 training  and 1,402 test videos and use these splits
for both tasks. Unsupervised VAD models must be trained only with normal data, 
so we extract precursor frames from each video for training. 
Supervised VAD and VAR models are trained using all training data.

\subsection{Task 1: Video Anomaly Detection (VAD)} \label{sec:VAD_eval}

\vspace{-5pt} \xhdr{Implementation Details.} We trained all ConvAE and ConvLSTMAE 
variants using AdaGrad with learning 
rate $0.01$ and batch size 24.
AnoPred, TAD, and their variants
are trained per the original papers. 
TAD+ML uses k-means ($k=10$) 
and center loss weight $1$~\cite{ionescu2019object,liu2019margin}. 
To train supervised methods, we first extract 
image features using ImageNet pre-trained 
ResNet50, then train each model with learning rate $0.0005$, batch 
size 16. All models are trained on NVIDIA 
TITAN XP GPUs. To fairly compare frame- 
and object-based methods,
we ignore videos with unknown category or 
without objects, resulting in 1,305 test videos.

\xhdr{Overall Results.} The top four rows of Table~\ref{tab:VAD} show performance of 
ConvAE and ConvLSTMAE with grayscale or optical flow inputs. Generally, using optical 
flow achieves better AUC, indicating motion is an  informative feature  for this task.
However, all  baselines achieve low STAUC, meaning that they cannot localize 
anomalous regions well. AnoPred achieves $67.5$ AUC 
but only $24.4$ STAUC, while  AnoPred+mask has $2.7$ lower AUC but $17.7$ higher STAUC. 
By applying instance masks, the model focuses on foreground objects 
to avoid computing high scores for background,
resulting in slightly lower AUC but much higher STAUC. This supports our hypothesis that \textit{higher AUC  does not imply a better VAD model, while STAUC reveals its ability to localize anomalous regions.} 
TAD outperforms 
AnoPred on both metrics by specifically focusing on object motion and location, both of 
which are important indicators of traffic anomalies. The margin learning (ML) module further improves TAD by a small margin. 
Our Ensemble method achieves the best AUC and STAUC among all methods, indicating 
that combining frame-level appearance and motion features is a direction worth investigating in future VAD research. 

State-of-the-art supervised methods such as TRN achieve higher AUC
than unsupervised methods. These methods focus on temporal modeling
and simplify spatial modeling by using pre-trained features.
We believe that exploring spatial modeling could further boost the 
performance of supervised methods. However, since these models directly 
predict an anomaly score for each frame rather than computing an anomaly score map, 
it is not straightforward to compute STAUC for them. Other ways such as a soft attention or class activation map might help  model explainability in the future~\cite{kim2018textual,zhou2016learning}. 

\begin{table}[t]
    \centering
    \scriptsize
    \caption{Benchmarks of VAD methods on the DoTA dataset. }
    \label{tab:VAD}
    \begin{tabular}{l@{\quad}@{\quad}l@{\quad}@{\quad}l@{\quad}@{\quad}r@{\quad}@{\quad}r@{\quad}}
        \toprule
        Method & Type & Input & AUC $\uparrow$ & STAUC $\uparrow$ \\ 
        \midrule
        ConvAE (gray)~\cite{hasan2016learning} & \multirow{8}{*}{Unsupervised} & Gray &  64.3 & 7.4 \\
        ConvAE (flow)~\cite{hasan2016learning}&  & Flow & 66.3 & 7.9  \\
        ConvLSTMAE (gray)~\cite{chong2017abnormal} &  & Gray & 53.8& 12.7 \\
        ConvLSTMAE (flow)~\cite{chong2017abnormal} &  & Flow & 62.5 & 12.2  \\
        AnoPred~\cite{liu2018future} &  & RGB & 67.5 & 24.4 \\
        AnoPred~\cite{liu2018future} + Mask &  & Masked RGB & 64.8 & 42.1 \\
        TAD~\cite{yao2019unsupervised} &  & Box + Flow & 69.2 & 43.3 \\
        TAD~\cite{yao2019unsupervised} + ML~\cite{ionescu2019object,liu2019margin} &  & Box + Flow & 69.7 & 43.7 \\ 
        Ensemble &  & RGB + Box + Flow & \textbf{73.0} & \textbf{48.5} \\
        \midrule
        FC & \multirow{4}{*}{Supervised} & \multirow{4}{*}{RGB} & 61.7 & -  \\ 
        LSTM~\cite{hochreiter1997long} & & & 63.7 & - \\
        Encoder-Decoder~\cite{cho2014learning} & & & 73.6 & -  \\
        TRN~\cite{xu2019temporal} & & & \textbf{78.0} & - \\
        \bottomrule
    \end{tabular}
\end{table}

\xhdr{Per-class Results.} Table.~\ref{tab:per_cls_VAD} shows per-class results of 
unsupervised methods: AnoPred, AnoPred+Mask, TAD+ML
and Ensemble. 
We observe that STAUC (unlike AUC) distinguishes performance by anomaly type, offering guidance as researchers seek to improve their methods.
For example, Ensemble has comparable AUCs on OC and VP anomalies ($73.4$ vs $70.1$) but significantly different STAUCs ($56.6$ vs $35.2$), 
showing that anomalous region localization is harder on VP. 
Similar trends exist for the AH*, LA*, VP* and VO* columns. 
Second, frame-level  and object-centric methods 
compensate each other in VAD as shown by the Ensemble method's highest AUC and STAUC values in most columns.
Third, localizing anomalous regions in non-ego anomalies is more difficult, as STAUCs on ego-involved anomalies are generally higher.
One reason is that ego-involved anomalies have better dashcam visibility and larger anomalous regions, making them easier to detect. 
Table~\ref{tab:per_cls_VAD} also shows the difficulties of detecting different categories, with
AH*, VP, VP*, VO* and  LA* especially challenging for all methods. We
observed that pedestrians in VP and VP* videos
become occluded or disappear quickly after an anomaly happens,
making it hard to detect the full anomaly event. 
AH* has a similar issue since sometimes the  vehicle ahead is 
largely occluded by the vehicle it impacts.
VO* is a rarer case in which a vehicle hits obstacles such as
bumpers or traffic cones which are typically not 
detected and are sometimes occluded by the anomalous vehicle.
Vehicles involved in LA* usually move towards each other slowly 
until they collide and stop, making the anomaly subtle
thus hard to distinguish.

\begin{table}[t]
    \centering
    \caption{Evaluation metrics of each individual anomaly class. Ego-involved and non-ego (*) anomalies are shown separately. VO and OO columns are not shown because they do not contain anomalous traffic participants.} 
    \label{tab:per_cls_VAD}
    \scriptsize
    \begin{tabular}{@{\;}l@{\quad}cccccccccccccc} 
        \toprule
        \multirow{1}{*}{} Method  & ST & AH & LA & OC & TC & VP & ST* & AH* & LA* & OC* & TC* & VP* & VO* & OO* \\ 
\midrule
        &\multicolumn{14}{c}{\textbf{Individual Anomaly Class AUC:}} \\
        AnoPred  & 69.9 & 73.6 & \textbf{75.2} & 69.7 & 73.5 & 66.3 & 70.9 & 62.6 & 60.1 & 65.6 & 65.4 & 64.9 & 64.2 & 57.8 \\ 
        AnoPred+Mask  & 66.3 & 72.2 & 64.2 & 65.4 & 65.6 & 66.6 & 72.9 & 63.7 &  60.6 & 66.9 & 65.7 & 64.0 & 58.8 & 59.9 \\ 
        TAD+ML & 67.3 & 77.4 & 71.1 & 68.6 & 69.2 & 65.1 & 75.1 & 66.2 & 66.8 & 74.1 & 72.0 & 69.7 & 63.8 & 69.2 \\ 
        Ensemble & \textbf{73.3} & \textbf{81.2} & 74.0 & \textbf{73.4} & \textbf{75.1} & \textbf{70.1} & \textbf{77.5} & \textbf{69.8} & \textbf{68.1} & \textbf{76.7} & \textbf{73.9} & \textbf{71.2} & \textbf{65.2} & \textbf{69.6} \\
        \midrule
         & \multicolumn{14}{c}{\textbf{Individual Anomaly Class STAUC:}}\\
        AnoPred & 37.4 & 31.5 & 32.8 & 34.3 & 33.6 & 24.9 & 25.9 & 15.0 & 12.5 & 13.0 & 20.9 & 14.0 & 8.2 & 8.8 \\ 
        AnoPred+Mask  & 51.8 & 51.9 & 45.1 & 50.3 & 47.5 & \textbf{41.0} & 45.3 & 31.1 & 33.8 & 42.5 & 40.3 & 25.3 & 22.9 & 33.8 \\ 
        TAD+ML & 47.4 & 55.6 & 46.3 & 52.2 & 47.2 & 26.6 & 45.1 & 33.6 & 38.5 & 46.9 & 39.3 & 25.6 & \textbf{29.0} & \textbf{44.4} \\
        Ensemble & \textbf{54.4} & \textbf{60.3} & \textbf{53.8} & \textbf{56.5} & \textbf{54.9} & 35.2 & \textbf{52.4} & \textbf{36.4} & \textbf{40.8} & \textbf{51.9} & \textbf{44.7} & \textbf{28.6} & 28.6 & 43.5 \\
        \bottomrule
    \end{tabular}
\end{table}

\begin{figure}[t]
    \centering
    \includegraphics[width=1.0\textwidth]{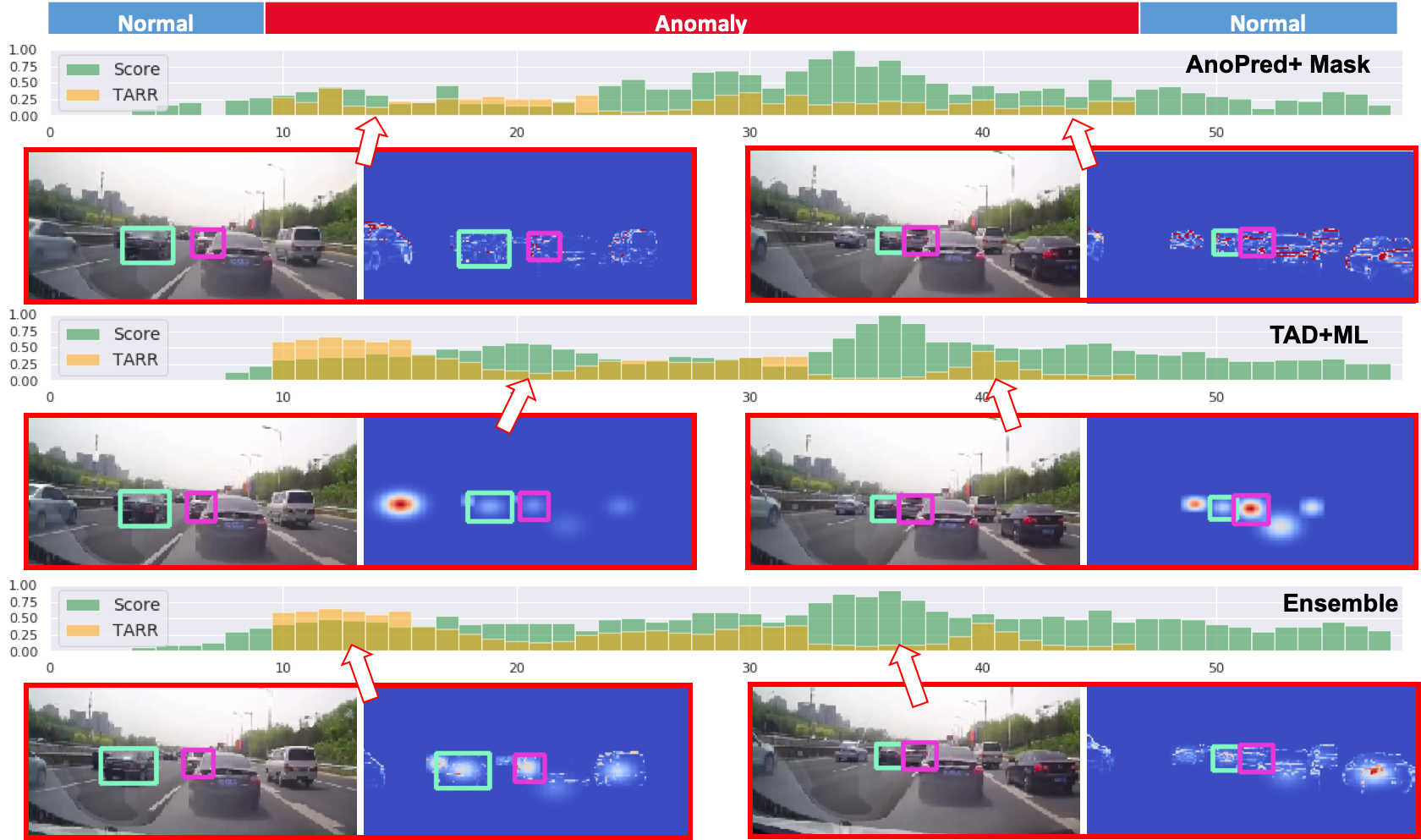}
    \caption{Per-frame anomaly scores and TARRs of three methods. Selected RGB frame and score maps are shown. Note that TARR only exists in positive frames.}
    \label{fig:vad_qualitative}
\end{figure}

\xhdr{Qualitative Results.} Fig.~\ref{fig:vad_qualitative} shows 
per-frame anomaly scores and $TARR$s
of three methods on a video where they all 
achieve high AUCs. 
AnoPred+Mask has low $TARR$ along the 
video, indicating failure of correctly 
localizing anomalous regions. 
TAD+ML computes high anomaly scores but 
low $TARR$ in the left example due to inaccurate 
trajectory prediction for the left car. 
In the right image, it finds one of the 
anomalous cars but also marks an unrelated 
car by mistake. 
Ensemble combines the benefits of 
both with anomaly scores for 
20-30th anomaly frames always higher than normal 
frames. It computes high TARR during 
10-20th anomaly frames as shown in the left score map.
The right map shows a failure case 
combining the failure of AnoPred+Mask and TAD+ML. 
Although these methods achieve high AUC, 
their spatial localization is limited 
per $TARR$. 
More qualitative results are 
shown in our supplement.

\subsection{Task 2: Video Action Recognition (VAR)} \label{sec:VAR_eval}
The goal of VAR is to assign each video clip to one anomaly category. We benchmark seven VAR methods on DoTA:
C3D~\cite{tran2015learning}, 
I3D~\cite{carreira2017quo}, 
R3D~\cite{tran2018closer}, 
MC3~\cite{tran2018closer},  
R(2+1)D~\cite{tran2018closer}, 
TSN~\cite{wang2016temporal} and 
SlowFast~\cite{feichtenhofer2019slowfast}. 
The previous training/test 
split is used.
Unknown UK(*) anomalies are ignored, 
yielding 3216 training and 1369 
test videos. 
We trained all 
models with SGD, learning rate 0.01 
and batch size 16 on NVIDIA TITAN XP GPUs. 
Models are initialized with 
Sports-1M~\cite{KarpathyCVPR14} (C3D) or 
Kinetics~\cite{kay2017kinetics} (rest) 
pre-trained weights; 0.5 probability random horizontal flip offers data augmentation. 
For evaluation, we randomly select ten clips from each test video per \cite{feichtenhofer2019slowfast} except TSN which uses 25 frames per video.

Table~\ref{tab:var_results} lists the backbone network of each 
model and its per-class accuracy. Although newer methods R(2+1)D and SlowFast 
achieve higher average accuracy, all candidates suffer from
low accuracy on DoTA, indicating that traffic anomaly classification 
is challenging. First, distant anomalies 
and occluded objects have low visibility thus are hard to classify. For example, VO(*) are hard to classify due to low visibility and diverse obstacle types per Section~\ref{sec:VAD_eval}. AH* and OC* are also difficult since the front or oncoming vehicles are often occluded.
Second, some anomalies are visually similar to others. 
For example, ST(*) are rare and look similar to AH(*) or LA(*) 
(Fig.\ref{fig:dota_sample}) since the only difference is whether the collided vehicle is starting, stopping, or stationary.
Third, anomaly category 
is usually determined by the frames around anomaly start time, while the later frames do not reveal this category clearly. 
We have observed $2$-$4\%$ accuracy improvement when testing models only on first half of each clip.
Additional benchmarks are available in our supplement.

\begin{table}[t]
    \centering
    \caption{VAR method per-class and mean top-1 accuracy with the DoTA dataset.}
    \label{tab:var_results}
    \tiny
    \begin{tabular}{llrrrrrrrrrrrrrrrrr}
        \toprule
          & & \multicolumn{16}{c}{Anomaly Class} &  \\ \cmidrule{3-18}
        Method & backbone &ST & AH & LA & OC & TC & VP & VO & OO & ST* & AH* & LA* & OC* & TC* & VP* & VO* & OO* & AVG \\ 
        \midrule
        TSN & ResNet50 & 18.2 & 67.2 & \textbf{52.9} & \textbf{53.8} & \textbf{71.0} & 0.0& 0.0& 61.6 & 0.0 & 14.7 & 25.3 & 6.7 & 48.1 & 9.5 & 0.0 & \textbf{53.4} & 30.2 \\ 
        C3D & VGG16 &25.5 & 61.8 & 43.9 & 47.8 & 57.9 & 3.3 & 4.4 & 52.9 & 1.2 & 18.4 & 36.0 & 6.7 & 55.9 & 8.6 & 6.0 & 33.2 & 29.0\\ 
        I3D & InceptionV1 & 10.0 & 62.4 & 45.8 & 45.8 & 62.2 & 2.8 & 6.9 & 66.6 & 2.4 & \textbf{28.1} & 24.5 & 4.7 & 60.3 & 9.5 & 5.0 & 37.6 & 29.7\\ 
        R3D & ResNet18 & 0.0 & 56.5 & 49.6 & 49.8 & 66.6 & 4.4 & 6.2 & 47.7 & 1.8 & 17.6 & 32.2 & 1.0 & 48.3 & \textbf{15.2} & 6.5 & 48.0 & 28.2 \\  
        MC3D & ResNet18 & 6.4 & 62.9 & 40.1 & 57.7 & 64.5 & 16.7 & 0.0 & 61.5 & 2.4 & 18.1 & 20.2 & 4.0 & 62.2 & 4.8 & \textbf{6.5} & 45.6 & 29.6 \\ 
        R(2+1)D & ResNet18 & 4.5 & 64.7 & 42.8 & 47.6 & 68.7 & \textbf{25.6} & 5.6 & 64.4 & \textbf{9.4} & 14.3 & 24.3 & 2.3 & \textbf{64.7} & 9.5 & 0.0 & 47.8 & \textbf{31.0} \\ 
        SlowFast & ResNet50 & 0.0 & \textbf{70.0} & 46.0 & 48.9 & 67.2 & 5.6 & \textbf{13.1} & \textbf{68.3} & 5.9 & 24.9 & \textbf{37.2} & 3.3 & 64.0 & 0.0 & 0.0 & 41.3 & \textbf{31.0} \\
        \bottomrule
    \end{tabular}
\end{table}

\section{Conclusion and Future Work}

This paper investigated a \textbf{When}-\textbf{When}-\textbf{What} pipeline for traffic anomaly detection. 
We introduced a large-scale dataset containing temporal, spatial, and categorical
annotations and benchmarked state-of-the-art VAD and VAR methods. 
We proposed a new spatial-temporal area under curve (STAUC) metric to better evaluate 
VAD performance. Experiments showed STAUC outperforms AUC but that traffic video 
anomaly detection and classification problems are far from solved. 
DoTA offers the community new data for further VAD and VAR research and also can be used to study important object (visual saliency) detection, online detection of traffic anomaly, and validation and verification of autonomous driving efforts.

\section{Acknowledgement}
This research has been supported by the National Science Foundation
under awards CNS 1544844 and CAREER IIS-1253549,
and by the IU Office of the Vice Provost for Research,
the IU College of Arts and Sciences, and the IU Luddy School of Informatics,
Computing, and Engineering through the Emerging Areas of Research Project ``Learning: Brains, Machines, and Children.''
We also thank Derek Lukacs and RedBrickAI\footnote{https://www.redbrickai.com/} for supporting our 
data annotation work. 
The views and conclusions contained in this
paper are those of the authors and should not be interpreted
as representing the official policies, either expressly or implied,
of the U.S. Government, or any sponsor.

\bibliographystyle{splncs04}
\bibliography{Reference}

\pagestyle{headings}

\title{When, Where, and What? A New Dataset for \\ Anomaly Detection in Driving Videos \\ -- Supplementary Material} 
\authorrunning{Yao~\etal} 
\author{Yu Yao$^{1}$
\hspace{1cm} Xizi Wang$^2$
\hspace{1cm} Mingze Xu$^3\thanks{This work was done while the author was at Indiana University.}$
\hspace{1cm} Zelin Pu$^1$\\
\hspace{1cm} Ella M. Atkins$^{1}$
\hspace{1cm} David J. Crandall$^{2}$ 
\\
$^{1}$University of Michigan
\hspace{0.45cm} $^{2}$Indiana University
\hspace{0.45cm} $^{3}$Amazon Rekognition  \\
{\tt\small \{brianyao,ematkins\}@umich.edu, \{xiziwang,djcran\}@indiana.edu}
}
\institute{}
\titlerunning{When, Where, and What? 
Anomaly Detection in Driving Videos}

\maketitle

\section{Additional DoTA Dataset Example}
\begin{figure}[htbp]
    \centering
    \includegraphics[width=0.8\textwidth]{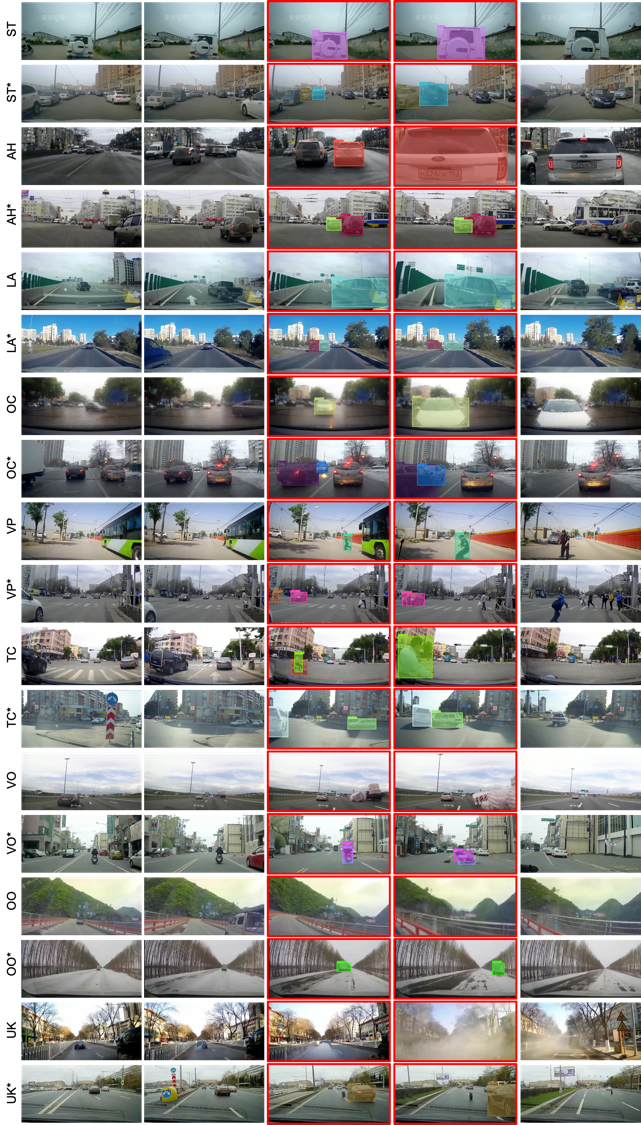}
    \caption{Sampled sequences from the DoTA dataset. Frames with 
    a red boundary are anomalous frames. Spatial annotations are 
    shown as shadowed bounding boxes. Short anomaly category labels 
    with * indicate non-ego anomalies.}
    \label{fig:extra_dota_example}
\end{figure}

Fig.~\ref{fig:extra_dota_example} shows one sampled frame sequence 
for each anomaly category in our DoTA dataset. Each row shows five 
frames sampled from one DoTA video, including two frames from the 
normal precursor, two frames from the anomaly window (marked by a 
red boundary), and one frame from the post-anomaly. The annotated 
bounding boxes of anomalous objects are shown by shadowed rectangles, 
and objects across frames are in consistent colors. Anomaly 
category abbreviations are listed to the left, where ``*" indicating 
non-ego anomalies.

This figure illustrates that some samples from different categories 
look similar, for example ST (row 1) is similar to both AH (row 3) 
and OC (row 7) except that in ST the front car is stationary. 
The AH* sample is similar to the OC* sample since it is difficult 
to distinguish front and rear vehicle views. The VP sample is close 
to the TC sample due to the similarity between a pedestrian and a 
rider. Moreover, some non-ego anomalies can have low visibility due 
to their distance from the camera, such as the VP* and the OO* 
example in Fig.~\ref{fig:extra_dota_example}. VO and VO* are 
anomalies where vehicles collide with unexpected/auxiliary 
obstacles such as dropped cargo and traffic cones. Note that VO 
and OO are two anomaly categories with no bounding box label 
typically provided; by definition, VO and OO do not involve traffic 
participants.

\section{Additional VAD (Task 1) Results} 
 
\begin{figure}[htbp]
    \centering
    \begin{subfigure}[t]{0.9\textwidth}
        \centering
        \includegraphics[width=1.0\textwidth]{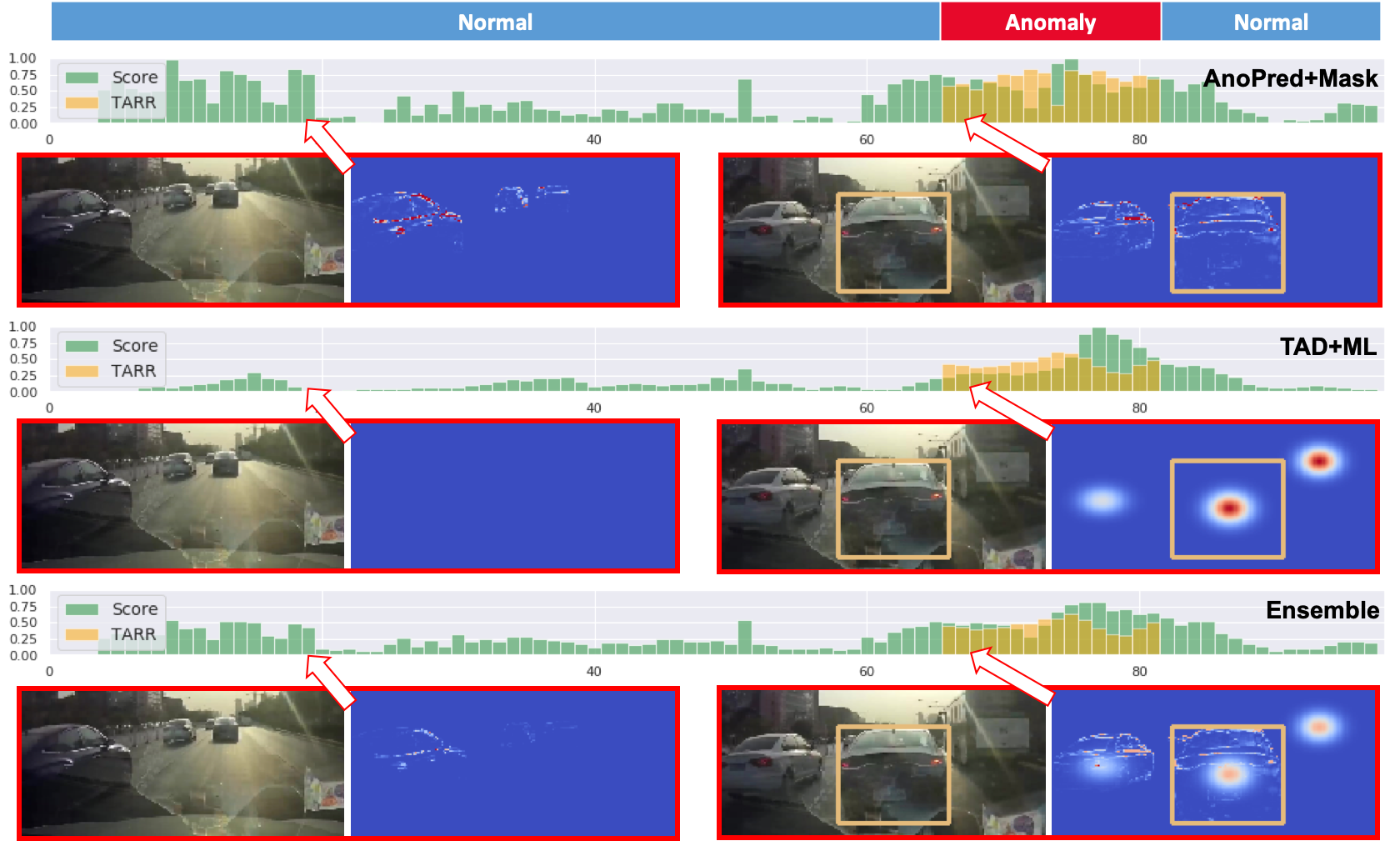}
        \caption{An example with Ensemble method outperforming each individual method.}
        \label{fig:qualitative1}
    \end{subfigure}
    \vspace{15pt}
    \begin{subfigure}[t]{0.9\textwidth}
        \centering
        \includegraphics[width=1.0\textwidth]{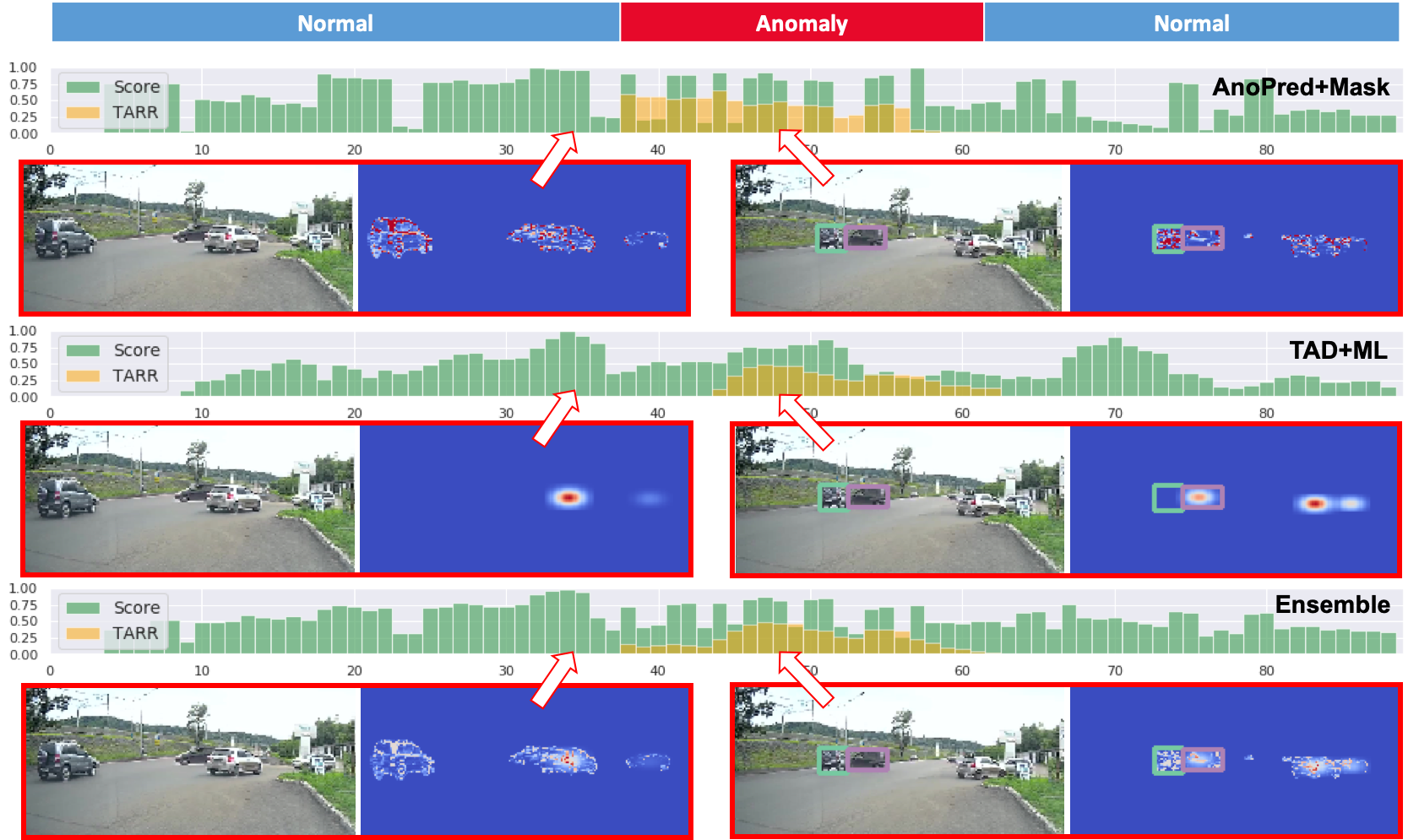}
        \caption{A failure case where both spatial and temporal performance is bad.}
        \label{fig:qualitative2}
    \end{subfigure}
    \caption{Additional qualitative results.}
    \label{fig:extra_qualitative}
\end{figure}

We present more qualitative results of AnoPred+Mask, TAD+ML and 
Ensemble methods in this supplement section. 
Fig.~\ref{fig:qualitative1} shows an ego-involved ahead collision 
(AH). AnoPred+Mask computes a high anomaly score in the early 
frames by mistake since the prediction of the left car is 
inaccurate, as shown in the score map. TAD+ML computes a low 
anomaly score for this frame and therefore the Ensemble method 
benefits. The right example shows the TAD+ML method correctly 
computing a high score for the ahead car but also another high 
score for the bus on the right. The ensemble benefits from 
AnoPred+Mask so that it focuses more attention on the ahead car 
instead of the bus. Fig.~\ref{fig:qualitative2} shows a failure 
case where all methods perform poorly in detecting a non-ego 
turning/crossing accident (TC*). The left example shows that all 
methods compute high anomaly scores for normal frames, where the 
silver car had to brake before turning to the right to avoid the 
black car which is turning into its lane. This example reveals 
that the tested unsupervised methods predict false alarms for 
near-incidence events. The right example shows that TAD+ML misses 
one of the anomalous cars, which is captured by AnoPred+Mask. 
This can be caused by the failure of object tracking in collision 
scenarios.

\section{Additional VAR (Task 2) Results}
\begin{figure}[t]
    \centering
    \begin{subfigure}[t]{0.45\textwidth}
        \centering
        \includegraphics[width=1.0\textwidth]{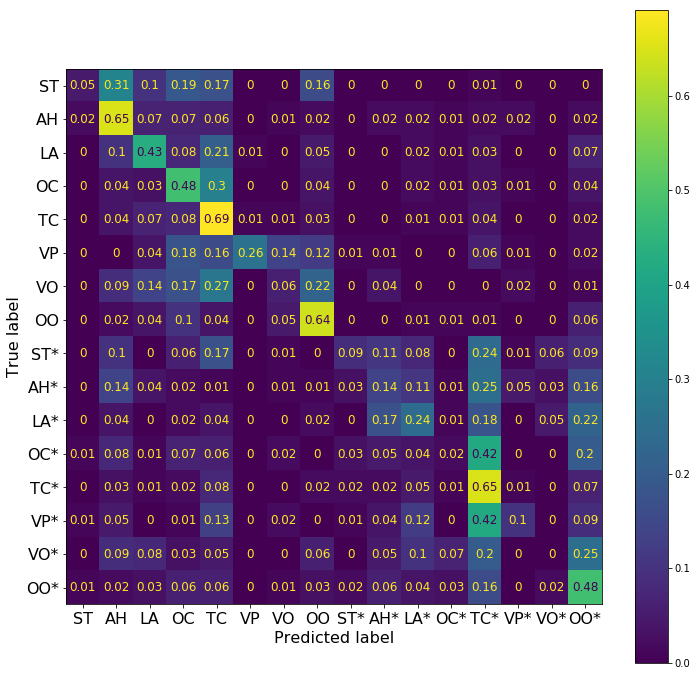}
        \caption{R(2+1)D}
        \label{fig:r2plus1d}
    \end{subfigure}
    ~
    \begin{subfigure}[t]{0.45\textwidth}
        \centering
        \includegraphics[width=1.0\textwidth]{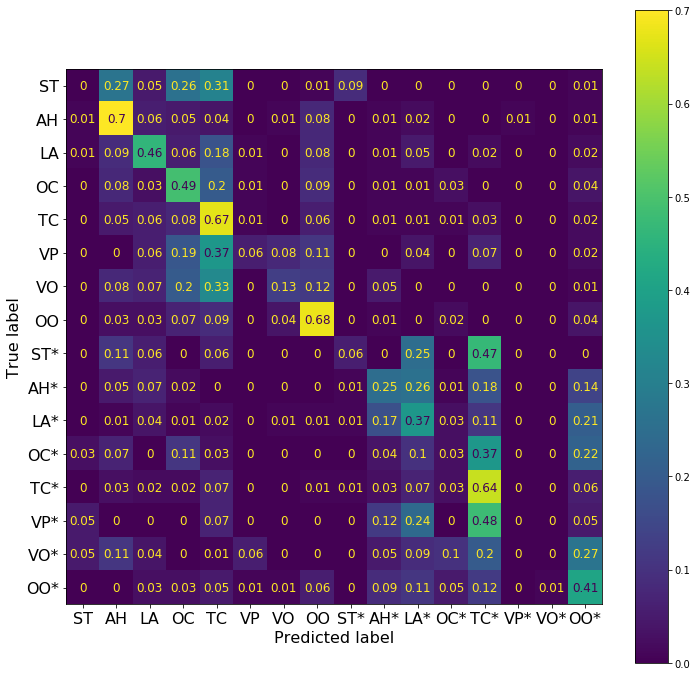}
        \caption{SlowFast}
        \label{fig:slowfast}
    \end{subfigure}
    \caption{Confusion matrix of two state-of-the-art VAR methods 
    on DoTA.}
    \label{fig:conf_matrices}
\end{figure}
In our submitted paper, we benchmarked several state-of-the-art video 
action recognition (VAR) models on DoTA dataset. 
Fig.~\ref{fig:conf_matrices} show the confusion matrices of R(2+1)D 
and SlowFast, two of the best models evaluated in our experiments. 
In addition to Table 5 in the paper, the confusion matrix shows the 
most confusing categories to help us understand challenging 
scenarios provided in the DoTA dataset. We make three observations 
from Fig.~\ref{fig:conf_matrices}. 
First, both models have similar confusion matrices, indicating that 
they perform similarly on DoTA dataset. 
Second, some categories are confused with other specific categories 
due to their similarities. 
Among all categories, TC, TC*, OC and OO* are four classes for which 
many categories are confused. One reason is that there are a large 
number of samples for these categories in DoTA. Another reason is 
the similarities among categories. For example OO* is usually an 
out-of-control vehicle swerving on the road and finally leaving 
the roadway. Other non-ego anomalies, while having their own 
features, often result in similar irregular motions, resulting 
in confusion with OO*. 
Third, ego-involved categories are usually not confused with 
non-ego categories. This indicates that although the per-class 
recognition is difficult, current methods could capably 
distinguish ego-involved and non-ego anomalies. 

\begin{table}[t]
    \centering
    \caption{Online Video Action Detection on our DoTA dataset. 
    "*" indicates non-ego anomaly categories.}
    \label{tab:VAR}
    \resizebox{\textwidth}{!}{
    \begin{tabular}{l@{\quad}ccccccccccccccccc}
        \toprule
         & \multicolumn{16}{c}{Anomaly Category}  &  \\
        \cmidrule{2-17}
        Method & ST & AH & LA & OC & TC & VP & VO & OO & ST* & AH* & LA* & OC* & TC* & VP* & VO* & OO* & mAP\\
        \midrule
        FC  & \textbf{2.5} & 13.9 & 10.6 & 6.2 & 16.3 & 0.8 & \textbf{1.2} & 21.0 & 0.6 & 2.9 & 3.0 & 0.6 & 8.0 & \textbf{1.2} & 0.7 & 7.6 & 9.9 \\ 
        LSTM  & 0.6 & 19.9 & 15.1 & 9.2 & 25.3 & 2.4 & 0.6 & 34.3 & 0.6 & 3.8 & 5.0 & 1.5 & 11.0 & 1.2 & 0.5 & 13.3 & 12.9\\ 
        Encoder-Decoder & 0.5 & 20.1 & 15.6 & 10.4 & 28.1 & \textbf{2.9} & 0.7 & \textbf{39.9} & \textbf{0.8} & 3.7 & 7.4 & 2.5 & 14.7 & 1.2 & 0.5 & 13.2 & 14.5\\ 
        TRN & 1.0 & \textbf{22.8} & \textbf{20.6} & \textbf{15.5} & \textbf{30.0} & 1.5 & 0.7 & 32.3 & 0.7 & \textbf{4.0} & \textbf{10.2} & \textbf{2.9} & \textbf{17.0} & 1.2 & \textbf{0.7} & \textbf{13.8} & \textbf{15.3} \\ 
        \bottomrule
    \end{tabular}}
\end{table}

\section{Task 3: Online Action Detection}
We provide benchmarks for online video action detection on 
DoTA dataset. Online action detection recognizes the anomaly 
type by only observing the current and past frames, making it 
suitable for autonomous driving applications. Since online 
action detection does not have a full observation of the whole 
video sequence, online action detection is considered a more 
difficult task than is traditional VAR. In this supplementary 
material, we provide benchmarks of several state-of-the-art 
online action detection methods on DoTA dataset. We use the 
same four online methods that have been used in supervised 
VAD: \textbf{FC}, \textbf{LSTM}, \textbf{Encoder-decoder} 
and \textbf{TRN}. The only difference is that the classifiers 
are designed to predict only one out of the 16 anomaly categories. 
We use the same training configurations to train these models. 
Table~\ref{tab:VAR} shows the per-class average precision (AP) 
and the mean average prediction (mAP).

\xhdr{Quantitative Results.} We observe that although TRN, a 
state-of-the-art method, achieves the highest mAP, all methods 
suffer from low precision on DoTA. Similar to what we have 
observed in the paper's VAD and VAR experiments, online action 
detection is also difficult for ST, ST*, VP, VP*, VO and VO*. 
AH* an OC* are also difficult due to the highly occluded front 
of a typical oncoming vehicle. We also observe that ego-involved 
anomalies are easier to recognize than non-ego anomalies due to 
their higher visibility.

\begin{figure}[t]
    \centering
    \includegraphics[width=0.9\columnwidth]{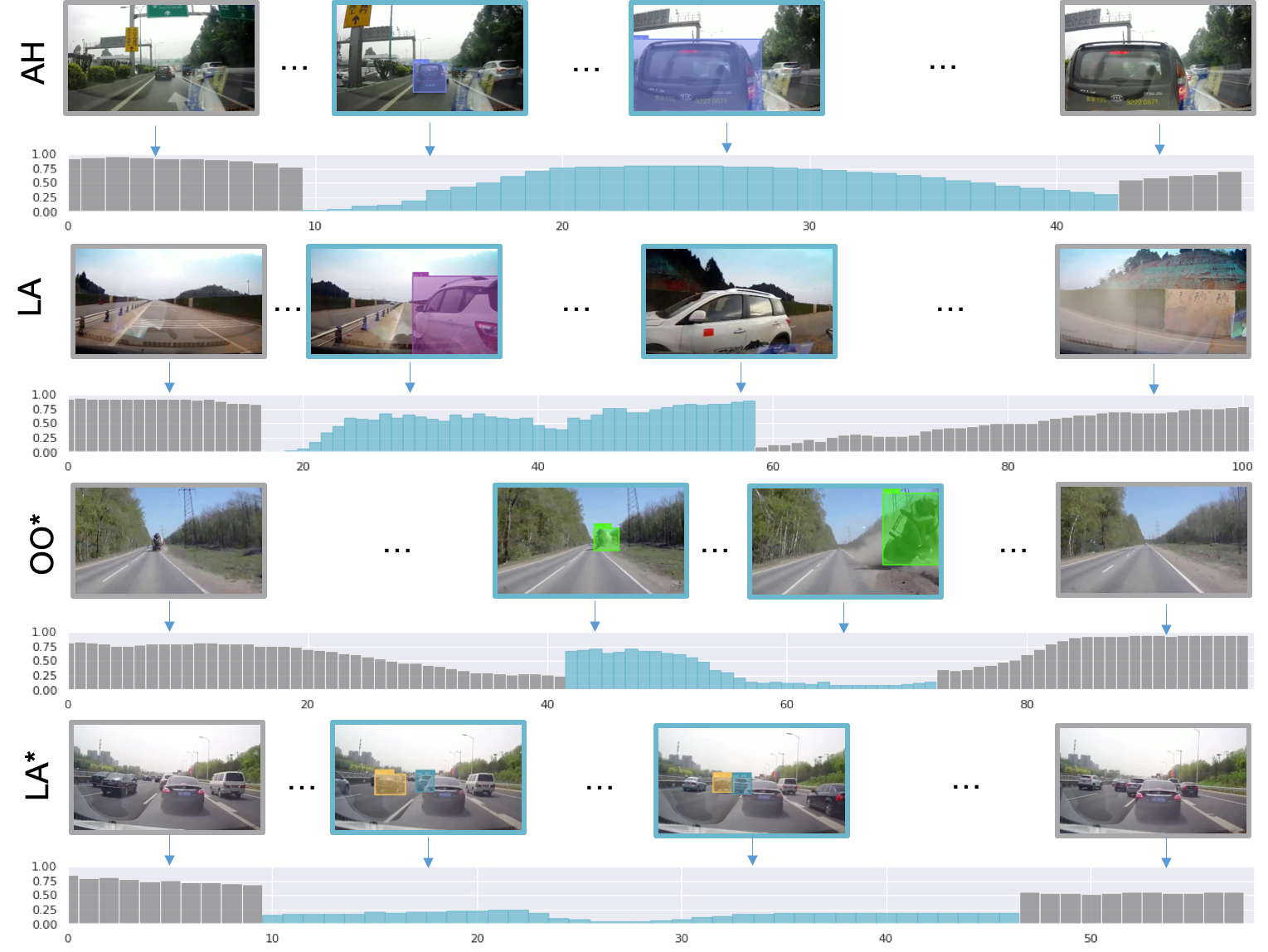}
    \caption{Qualitative results of Temporal Recurrent 
    Network (TRN) on our DoTA dataset. 
    The bar plots show classification confidences of 
    each video frame. Gray bars are confidences of 
    "background" (or "normal") classes while cyan bars 
    are confidences of ground truth anomaly classes. 
    The top two rows are two ego-involved anomalies, 
    while the 3rd row is a non-ego out-of-control anomaly. 
    The 4th row is a case where TRN fails to detect a lateral 
    collision.}
    \label{fig:online_var_qualitative}
\end{figure}

\xhdr{Qualitative Results.} Fig.~\ref{fig:online_var_qualitative} 
shows some examples of TRN results on our DoTA dataset. 
The bar plots show the classification confidences of each frame. 
Cyan colors represent anomalous frames while gray colors 
represent background (normal) frames. We make the following 
observations from this experiment: 1) Transition frames between 
normal and abnormal events are hard to classify. For example class 
confidences are low at the frames where color changes, i.e., 
anomaly start and end frames; 2) Subsequent frames after an 
anomaly begins can be hard to detect. For example confidence 
significantly decreases at around the 40th frame of first example 
and the 60th frame of the third example; 3) Visually similar 
anomalies and gentle anomalies are hard to detect. In the bottom 
failure case, the confidence of ground truth anomaly class 
LA* is always low. These frames are either classified as 
background (normal) or AH* due to the fact that this LA* 
anomaly is visually similar to a typical AH* anomaly since 
this collision is relatively gentle.

\end{document}